\documentclass[a4paper]{article}

\usepackage{INTERSPEECH2020}
\usepackage{times}
\usepackage{soul}
\usepackage{url}
\usepackage[hidelinks]{hyperref}
\usepackage[utf8]{inputenc}
\usepackage{caption}
\usepackage{graphicx}
\usepackage{amsmath}
\usepackage{amsthm}
\usepackage{booktabs}
\usepackage{algorithm}
\usepackage{algorithmic}
\usepackage{subcaption}
\usepackage{tikz}
 \usetikzlibrary{positioning}
\usepackage{multirow}
\title{Sequential End-to-End Intent and Slot Label Classification and Localization}
\name{Yiran Cao$^1$, Nihal Potdar$^1$, Anderson R. Avila$^2$}
\address{
  $^1$University of Waterloo\\
  $^2$Huawei Noah's Ark Lab}
\email{\{yiran.cao, nihal.potdar\}@waterloo.edu, anderson.avila@huawei.com}

\begin{document}

\maketitle
\begin{abstract}
Human-computer interaction (HCI) is significantly impacted by delayed responses from a spoken dialogue system. Hence, end-to-end (e2e) spoken language understanding (SLU) solutions have recently been proposed to decrease latency. Such approaches allow for the extraction of semantic information directly from the speech signal, thus bypassing the need for a transcript from an automatic speech recognition (ASR) system. In this paper, we propose a compact e2e SLU architecture for streaming scenarios, where chunks of the speech signal are processed continuously to predict intent and slot values. Our model is based on a 3D convolutional neural network (3D-CNN) and a unidirectional long short-term memory (LSTM). We compare the performance of two alignment-free losses: the connectionist temporal classification (CTC) method and its adapted version, namely connectionist temporal localization (CTL). The latter performs not only the classification but also localization of sequential audio events. The proposed solution is evaluated on the Fluent Speech Command dataset and results show our model ability to process incoming speech signal, reaching accuracy as high as 98.97 \% for CTC and 98.78 \% for CTL on single-label classification, and as high as 95.69 \% for CTC and 95.28 \% for CTL on two-label prediction.

\end{abstract}

\section{Introduction}

Spoken language understanding (SLU) aims at extracting structured semantic representations, such as intent and slots, from the speech signal \cite{mhiri2020low}. These representations are crucial to enable speech as the primary mode of human-computer interaction (HCI) \cite{he2019streaming}. Traditional SLU solutions rely on the text transcription generated by an automatic speech recognition (ASR) module, followed by a natural language understanding (NLU) system, responsible for extracting semantics from the ASR output \cite{qian2019spoken}. As described in \cite{shivakumar2019incremental}, in such scenarios the ASR typically operates on chunks of the incoming speech signal and outputs the transcript for each segment. The NLU then waits until all the speech segments are transcribed before processing the ASR output. This has significant latency implications. Another issue is that each module is trained and optimized separately. While the ASR optimization aims at minimizing word error rate, the NLU is often optimized on clean text with the assumption of error-less transcriptions from the ASR \cite{serdyuk2018towards}. Besides, this approach provides a cumulative error that propagates from each module, adding up to the overall SLU error.


Recently, we have witnessed an increasing interest in reducing the latency of the SLU task. Low-latency leads to more naturalness while interacting with a computer system and can ultimately improve the user experience (UX) \cite{porcheron2018voice}. To this end, a handful of studies have specifically addressed the problem. In \cite{shivakumar2019incremental}, the authors proposed a recurrent neural network (RNN) for processing the output of an ASR system in an online fashion. Their streaming SLU solution is based on an online NLU that processes word sequences of arbitrary length and incrementally provides multiple intent predictions. Similarly, the authors in \cite{liu2016joint} propose an RNN-based model that jointly performs online intent detection and slot filling as input word embeddings arrive. Results show that the joint training model provides high accuracy for intent detection and language modeling with a small degradation on slot filling compared to the independent training models. Although these approaches show reasonable performance, they rely on the strong assumption of error-less transcriptions from the ASR as their NLU system is often trained on clean text. Moreover, they can not be considered end-to-end (e2e) solutions as their models are based on the ASR transcription.

To mitigate this, other studies have proposed the extraction of semantic information directly from audio. For example, several e2e SLU encoder-decoder architectures are investigated in \cite{haghani2018audio}. The authors showed that better performance is achieved when an e2e SLU solution that performs domain, intent, and argument prediction is jointly trained with an e2e ASR model that learns to generate transcripts from the same input speech. Another recent study introduces the Fluent Speech Command (FSC) dataset \cite{lugosch2019speech}. The authors present a pre-training strategy for e2e SLU models. Their approach is based on using ASR targets, such as words and phonemes, that are used to pre-train the initial layers of their final model. These classifiers once trained are discarded and the embeddings from the pre-trained layers are used as features for the SLU task. Improved performance on large and small SLU training sets was achieved with the proposed pre-training approach. Similarly, in \cite{chen2018spoken}, the authors also proposed to fine-tune the lower layers of an end-to-end CNN-RNN based model that learns to predict graphemes. This pre-trained acoustic model is optimized with the connectionist temporal classification (CTC) loss and then combined with a semantic model to predict intents. 

The aforementioned research efforts have been either on developing online NLU or non-streamable e2e SLU. In the light of that, investigating a complete end-to-end low-latency streaming SLU solution is necessary. In this paper, we propose a compact e2e streamable SLU solution that (1) eliminates the need for an ASR module with (2) an online architecture that provides intent and slot predictions while processing incoming speech signals. To achieve that, a 3-dimensional convolutional neural network (3D-CNN) combined with a unidirectional long-short term memory (LSTM) is explored. We compare two alignment-free loss functions: the CTC method and its adaptation, namely the connectionist temporal localization (CTL) function. Both methods will be discussed in section \ref{sec:sequential}. We use the FSC dataset to perform our experiments and results show our model achieving accuracy as high as 98.97 \% for single intent+slot classification and 95.69 \% for multiple intent+slot detection.

The remainder of this paper is organized as follows. In Section \ref{sec:sequential}, we discuss sequential labeling. Section \ref{sec:model_arch} presents the proposed architecture. Section \ref{sec:setup} describes our experimental setup and Section \ref{sec:results} presents our experimental evaluation. Section \ref{sec:conclusion} gives the conclusion.

\begin{figure}
\centering
  \includegraphics[width=0.99\linewidth]{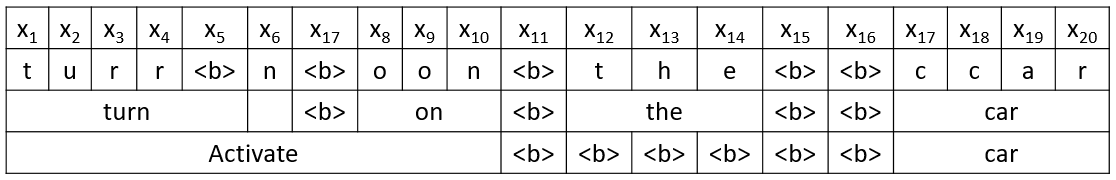}
  \caption{CTC alignment for $X = [x_1, x_2,..., x_n]$, output Y$_{asr}$  $=$ $[$ t,u,r,n,\_,o,n,\_,t,h,e,\_,c,a,r$]$ and Y$_{slu}$ = $[$activate, car$]$.}
  \label{fig:ctc}
\end{figure}

\section{Sequential Labeling}
\label{sec:sequential}

\subsection{Streaming Spoken Language Understanding}
\label{subsec:slu}

In a streaming e2e SLU scenario, given the input $X = [x_1, x_2,..., x_n]$ of acoustic features of length $N$ and the corresponding sequence of semantic outputs $Y = [y_1, y_2,..., y_u]$ of length $U$, the precise alignment of $X$ and $Y$ is not known and typically $N > U$. Unlikely ASR, for e2e SLU the gap between input and output length is higher as the semantic label prediction is conditioned to a larger input context. The period of silence in the audio tends to even increase this gap. The goal is to learn the distribution $P(Y|X,\theta)$. However, different from the non-streaming scenario, predictions are made for a given timestep, $t$, with the model incrementally predicting multiple output intents (or blank symbols for silence), while accounting for the context dependency from previous predictions. 

\subsection{Connectionist Temporal Classification}
\label{subsec:ctc}

The CTC method was first motivated to train RNNs of unsegmented data \cite{graves2006connectionist}. Previous to CTC, training RNNs required prior segmentation of the input sequence. For that, each input segment was labeled and the RNN was trained to predict an output for each segment at a time. This required a post-processing step to consolidate the output predictions into a sequence of labels \cite{graves2006connectionist}. With CTC, however, prior segmentation is no longer needed as the method allows a sequence-to-sequence mapping free of alignment. For acoustic modeling, for example, CTC automatically learns the alignments between the input sequence of acoustic frames and the respective sequence of output labels \cite{miao2016empirical}. As illustrated in Figure~\ref{fig:ctc}, CTC defines the so-called naive alignment by matching the input and output length adding the blank tokens ($<$b$>$), and repeating output predictions. After this, blank tokens are removed and repeated predictions are collapsed. 

Formally, the conditional probability of a single alignment (or path), $\alpha$, is the product of the probabilities of observing $\alpha_t$ at time $t$ and can be represented as

\begin{equation}
    P(\alpha|X, \theta) = \prod_{t=1}^{T} P(\alpha_t | X, \theta)
\end{equation}

\noindent where $\alpha_t$ represents a given label. Because $P(\alpha|X)$ defines mutually exclusive paths, the conditional probability for a sequence output is given by the sum of the probabilities of all paths corresponding to it: 
\begin{equation}
    P(Y|X, \theta) = \sum_{\alpha \in A_{X,Y}} \prod_{t=1}^T P(\alpha_t|X, \theta)
\end{equation}

\noindent where $A_{X,Y}$ is the set of all valid alignments. The CTC loss is then defined as 

\begin{equation}
    L_{CTC}(X,Y) = -log\sum_{\alpha \in A_{X,Y}} \prod_{t=1}^T P(\alpha_t|X, \theta)
\end{equation}

CTC considers no dependency between previous time steps which allows for frame-wise gradient propagation, but limits the possibility of learning sequential dependencies \cite{kamath2019deep}.









\begin{figure}
\centering
  \includegraphics[width=0.99\linewidth]{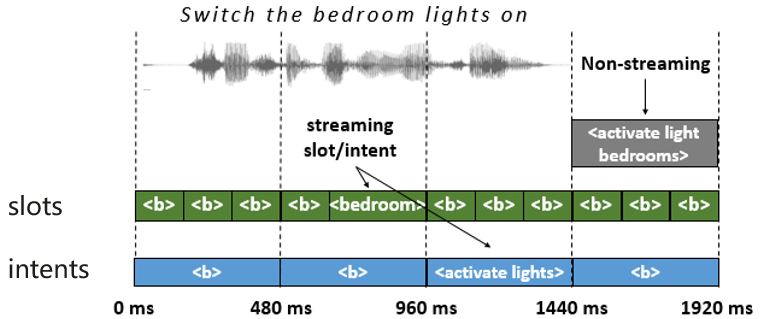}
  \caption{Streaming intent + slot approach versus a non-streaming one. Note that slot has a higher prediction rate than intent (3 times faster in our experiments) as it depends on less temporal speech context.}
  \label{fig:prediction}
\end{figure}

\subsection{Connectionist Temporal Localization}
\label{subsec:ctl}

CTL was first introduced in \cite{wang2019connectionist} for classification and localization of sound event occurrences in an audio stream. CTC has been previously applied for sound event detection (SED), however, it was found that the method presented the so-called "peak clustering," especially for long events \cite{wang2017first}. In such cases, onset and offset labels will result in a peak of the frame-probabilities. During training, because the adjacent onset and offset labels of long events occur next to each other, CTC may interpret them as the existence of boundaries instead of the existence of an event. Because the CTC loss function focus on predicting sequential labels in the correct order regardless of any temporal constraints, the recurrent layer will derive onset and offset labels next to each other as it minimizes memory effort \cite{wang2019connectionist}. Moreover, the model detects event boundaries which leads to high frame-probabilities surrounding onset and offset events and remains inactive for the period that the event is on, even when changes in the acoustic features are observed.

Three improvements are proposed to overcome the peak clustering issue \cite{wang2019connectionist}. First, boundary probabilities are attained from network event probabilities using a "rectified delta" operator. This assures that the network predicts
frame-wise probabilities of events and not
of the event boundaries, which leads to
different predictions for different acoustic features \cite{wang2019connectionist}. The event boundaries are calculated then as follows,

\begin{equation}
\begin{split}
    z_t(\acute{E}) = max[0, y_t(E)-y_{t-1}(E)] \\
    z_t(\grave{E}) = max[0, y_{t-1}(E)-y_t(E)]
\end{split}
\end{equation}

\noindent where $y_t(E)$ is the probability of the event $E$ at frame $t$, whereas $z_t(\acute{E})$ and $z_t(\grave{E})$ represent, respectively, onset and offset labels of event $E$ at frame $t$.

Second, the boundary probabilities at each frame are considered mutually independent, which allows the overlap of sound events. The independence assumption eliminates the need for a black symbol used by CTC to emit nothing at a frame, as well as to separate repetition of the same label \cite{wang2019connectionist}. 

The third modification implies that consecutive repeating labels are no longer collapsed. With these modifications, multiple labels can be emitted at the same frame, which can not be achieved with the standard CTC \cite{wang2019connectionist}. Thus the probability of emitting multiple labels at frame $t$ is attained as follow, 

\begin{equation}
    p_t(l_1,...,l_k) = \prod\nolimits_{i=1}^{k}z_t(l_i)\cdot \prod\nolimits_{l\not\in }[1-z_t(l_i)]
\end{equation}

\noindent with $p_t(l_1,...,l_k)$, being the probability of emitting the sequence of labels $L=(l_1,...,l_k)$ at frame $t$ without the need of temporal alignment. Thus, given the frame-probabilities of events $y_t(E)$ the probability of emitting the first i labels of L can be represented with the recurrence formula bellow \cite{wang2019connectionist}: 

\begin{equation}
    \alpha_t(i) = \sum_{j=0}^{i}\alpha_{(t-1)}(i-j) p_t(l_{i-j+1},...,l_i)
\end{equation}

\noindent where $\alpha_t(i)$ represents the probability of emitting the first $i$ labels at frame $t$ and $j$ is the number of labels emitted at that particular frame, i.e. zero, one or more labels. Initial values of $\alpha_0(i)$ are set to $1$ for $i = 0$. We refer the reader to \cite{wang2019connectionist} for more details regarding the CTL loss.

\begin{figure}
\centering
  \includegraphics[width=1\linewidth]{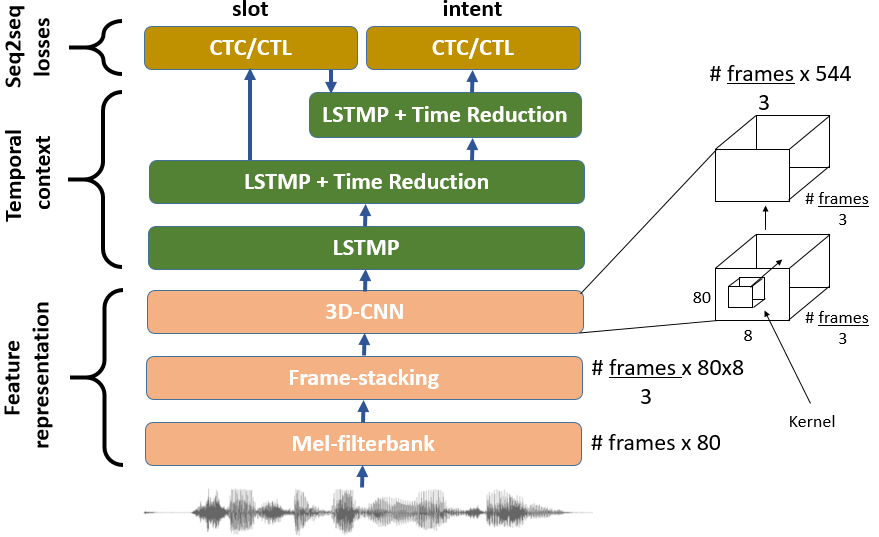}
  \caption{Diagram depicting the proposed end-to-end unidirectional RNN for mapping a sequence of speech frames into a sequence of slots+intents without predefined alignment.}
  \label{fig:model_architecture}
\end{figure}

\section{Neural Network Architecture}
\label{sec:model_arch}

The proposed model is depicted in Figure~\ref{fig:model_architecture} and is divided into three main parts. The first part focuses on learning feature representation from the speech signal. After extracting Mel-filterbank features, a stacking operation is performed. Specifically, we stacked 8 input frames with stride 3. The motivation for this is twofold. First, it aims at providing a larger set of context frames to the upper layers which help to alleviate the work of CTC and CTL as it reduces the number of output paths to be explored. Second, it minimizes latency as the frame rate is reduced. After the stacking step, a 3D-CNN is also applied. Adopted in other speech-related studies \cite{kim2017learning}\cite{ganapathy20183}\cite{meng2019speech}, our motivation is to learn better representation by processing time, frequency, and channel dimensions from the stacked Mel-filterbank features. Similar to the approach taken in \cite{ganapathy20183}, no padding was used and two 3D convolutional layers are used. Both layers have kernel size of 5x5x1 and stride of 2x2x1. 16 kernels are used in the first layer, followed by 32 in the second layer. As depicted in Figure~\ref{fig:model_architecture}, the temporal dynamics are preserved. Three long short-term memory (LSTM) layers are adopted to capture temporal context from the speech representation. Time reduction is applied only on the second and third LSTM layers along with a projection operation. The time reduction is performed by concatenating the hidden states of the LSTM by a factor of 4. While it results in fewer time steps, the feature dimension increases by the same factor. The feature dimension is controlled with a projection layer. Time reduction (or 'time convolution \cite{rao2017exploring}) is used to reduce the sequence length of encoded activation, also meant to minimize the gap between the length of input and output sequences. Note that the softmax from the second LSTM layer is used to predict slots, whereas the prediction of intents is based on the softmax from the third LSTM layer. Moreover, because intents and slots likely carry out related information, the output of the second layer (i.e. the prediction of slots) is used as additional input to the third layer.



\section{Experimental Setup}
\label{sec:setup}

\subsection{Datasets}

Two datasets are used in our experiments. To pretrain our models, we relied on the LibriSpeech corpus, which was introduced in \cite{panayotov2015librispeech}. Librispeech contains about 1000 hours of speech sampled at 16 kHz. Only the clean speech containing 360 hours of speech was used. The second dataset adopted was the Fluent Speech Commands dataset, which comprises single-channel audio clips sampled at 16 kHz. The data was collected using crowdsourcing, with participants requested to cite random phrases for each intent twice. It contains about 19 hours of speech, providing a total of 30.043 utterances cited by 97 different speakers. The data is split in such a way that the training set contains 14.7 hours of data, totaling 23,132 utterances from 77 speakers. Validation and test sets comprise 1.9 and 2.4 hours of speech, leading to 3,118 utterances from 10 speakers and 3,793 utterances from other 10 speakers, respectively. The dataset comprises a total of 31
unique intent labels resulted in a combination of three slots per audio: action, object, and location. The latter can be either “none”, “kitchen”, “bedroom”, “washroom”, “English”, “Chinese”, “Korean”, or “German”. More details about the dataset can be found in \cite{lugosch2019speech}. To simulate multi-intent scenarios, an additional version of the FSC dataset was generated. Namely FSC-M2, it is the result of concatenating two utterances from the same speaker into a single sentence. In this version, all the possible intent combinations were evenly distributed in the dataset. The training part, FSC-M2-Tr, contains 57,923 utterances, totaling approximately 74.27 hours, selected from the FSC training data, with the test part, FSC-M2-Tst, 
containing 8,538 utterances, roughly 11.80 hours from the FSC test data. 

\begin{table}
\centering
\caption{Experimental results on FSC for single-intent classification. Performance is reported in terms of accuracy (\%) for 31 intent targets.}
\scalebox{0.99}{
\begin{tabular}{lccc}
\toprule
Model & Intent \\
\midrule
RNN+Pre-training \cite{lugosch2019speech} &  98.80 \\
CNN+Segment pooling \cite{mhiri2020low} & 97.80 \\
CNN+GRU(SotA) \cite{tian2020improving} & 99.10\\
\midrule
3D-CNN+LSTM+CE & \textbf{99.26} \\
\bottomrule
\end{tabular}}
\label{tab:single_ce}
\end{table}

\subsection{Features}

\begin{table}
\centering
\caption{Results on multi-intent classification. The training set consists of 1 label and the testing set contains utterances with 1 and 2 labels, referred to as FSC-M1-Tst and FSC-M2-Tst.}
\scalebox{0.63}{
\begin{tabular}{lccccccc}
\toprule
&\multicolumn{3}{c}{FSC-M1-Tst}&&\multicolumn{3}{c}{FSC-M2-Tst}\\
\cmidrule{2-4}\cmidrule{6-8}
Model&Intent &Slot&Intent+Slot& &Intent&Slot& Intent+Slot\\
\midrule
CTC & 27.70 & 54.41 & 14.50 & & 29.92 & 24.83 & 5.30\\
+ Joint CE & 98.25 & 99.34 & 97.73 & & 53.00 & 60.55 & 41.31\\
+ Pretrained ASR & 99.15 & \textbf{99.76} & \textbf{98.97}  & & 56.04 & 64.21 & 44.30\\
\midrule
CTL & 12.76 & 55.54 & 3.60 & & 0.14 & 25.05 & 0.05\\
+ Joint CE & 89.05 & 95.35 & 85.68 & & 48.37 & 80.14 & 41.28\\
+ Pretrained ASR & 97.25 & 99.12 & 96.41 & & 44.50 & 81.22 & 39.29 \\ 
\midrule
CTL + MIL  & 98.54 & 98.39 & 96.99 && \textbf{78.21} & 86.00 & \textbf{69.10}\\
+ Pretrained ASR & \textbf{99.18} & 99.57 & 98.78 && 66.17 & \textbf{96.77} & 64.89\\
\bottomrule
\end{tabular}}
\label{tab:res_single}
\end{table}

In this work, audio signals are sampled at 16 kHz. As acoustic features, 80-dimensional log Mel-Filterbank features are adopted \cite{watanabe2018espnet}. To extract the Mel features, the audio signal is processed in frames of 320 samples (i.e., 20-ms window length), with a step size of 160 samples (that is, 10-ms hop-size). Global Cepstral Mean and Variance Normalization (CMVN) is applied, as a process that is commonly used with the aim to increase the robustness of ASR systems while mitigating the mismatch between training and testing data \cite{pujol2006real}\cite{ravanelli2019pytorch}

\subsection{Experimental Settings}

Our network was trained on mini-batches of 64 samples over a total of 200 epochs. In our experiments, we adopted learning rate of 0.0001 and dropout of 0.1. The optimizer used was Adam with weight decay of 0.2. We explored three strategies to optimize our model for the streaming scenario, including (1) using pure CTC or pure CTL as loss function; (2) using CTC or CTL jointly with CE. In the case of CTL, we also attempted to combine it with the multiple instance learning (MIL) technique \cite{wang2019connectionist}; and finally (3) adding a pretrained ASR to our model, optimized with the first layer with the CTC loss on character prediction. Note that CE and the MIL were only used during training. For the CE, we applied softmax to the last timestep and average the CTC and CE losses with a fixed weight of 0.6 and 0.4, respectively. Because the MIL is based on frame-wise probabilities just as CTL, their combination is straight-forward and consists of a simple weighted average of the two losses. The main difference is that the MIL aggregates the frame-probabilities into recording-level probabilities with a linear softmax pooling function \cite{wang2019connectionist}. Note that to ensure the streaming ability of our model, at the testing time only CTC or CTL are used. The third strategy consisted of pretraining an ASR model with the CTC loss. This procedure aimed at leveraging pre-trained embeddings by learning better representations from a large corpus. The model was trained to predict characters and only the first layer of our neural network was used. After training the ASR for 150 epochs, its weights were frozen and used with the entire recurrent neural network. In our experiments, we defined intent as the combination of action and object, which led to a total of 15 different intent labels. Location was defined as slot, which led to a total of 8 different slot labels. Our model was evaluated on 3 tasks: intent prediction; slot prediction; and intent+slot prediction. While intent predictions are attained using the outputs of the third layer, slot predictions rely on the output of the second layer, as illustrated in Figure~\ref{sec:model_arch}.  
 
\section{Experimental Evaluation}
\label{sec:results}

The performance of the proposed architecture is investigated in three different experiments. In the first one, we evaluate our model in a non-streaming scenario. The model is based on the pre-trained ASR and the architecture presented in Section\ref{sec:model_arch}. Only CE loss is used for optimizing our model. This experiment aims to compare our network with non-streamable e2e SLU solutions proposed in the literature. Results are reported in Table~\ref{tab:single_ce} and it shows our model achieving high accuracy in the three tasks and outperforming the other three baselines.

In the second experiment, we tackle a more challenging setting and our model is evaluated under the streaming regime. As described in Table~\ref{tab:res_single}, the model is trained with a single label (i.e. intent, slot, or intent+slot) and tested on one (FSC-M1-Tst) and two labels (FSC-M2-Tst). Results show that using the CE loss during training is a key strategy to boost the performance of both CTC and CTL. The use of a pretrained ASR is also crucial for achieving high accuracy. Also, combining CTL with MIL yields better accuracy overall. Identifying intent and slot correctly at the same time is difficult and provides lower accuracy when compared to the single task (i.e. single intent or single slot). Detecting 2 intents have a detrimental impact on the performance of our model. To overcome this, in the last experiment, our model is trained on two labels and tested on one and two labels as well. Results are presented on Table~\ref{tab:res_multi}. We can observe that training on two labels benefits the performance of our model. Although CTC and CTL present comparable results, one has to consider that CTL has the potential for performing localization, whereas CTC is limited to predicting a sequence. Moreover, CTL allows for overlap events.

\begin{table}
\centering
\caption{Results on multi-intent classification. The training set consists of 2 labels and the testing set contains utterances with 1 and 2 labels, referred to as FSC-M1-Tst and FSC-M2-Tst.}
\scalebox{0.63}{
\begin{tabular}{lccccccc}
\toprule
&\multicolumn{3}{c}{FSC-M1-Tst}&&\multicolumn{3}{c}{FSC-M2-Tst}\\
\cmidrule{2-4}\cmidrule{6-8}
Model&Intent &Slot&Intent+Slot& &Intent&Slot& Intent+Slot\\
\midrule
CTC & 22.83 & 0.00 & 0.00 & & 22.43 & 25.03 & 4.24\\
+ Joint CE & 97.49 & 99.23 & 96.88 & & 95.16 & 97.77 & 93.73\\
+ Pretrained ASR & 98.54 & 99.39 & 98.15  & & \textbf{96.45} & 98.47 & \textbf{95.69}\\
\midrule
CTL & 11.83 & 55.54 & 2.66 & & 1.17 & 26.58 & 0.20\\
+ Joint CE & 86.79 & 94.93 & 83.15 & & 76.45 & 88.43 & 69.31\\
+ Pretrained ASR& 95.38 & 99.10 & 94.67 & & 87.58 & 97.33 & 85.50 \\ 
\midrule
CTL + MIL & 96.70 & 98.89 & 95.75 && 93.14 & 96.83 & 90.30 \\
+ Pretrained ASR & \textbf{98.65} & \textbf{99.68} & \textbf{98.36} && 95.78 & \textbf{99.50} & 95.28 \\
\bottomrule
\end{tabular}}
\label{tab:res_multi}
\end{table}

\section{Conclusion}
\label{sec:conclusion}

In this paper, we evaluated a compact spoken language understanding (SLU) model optimized with two alignment-free losses: the connectionist temporal classification (CTC) and the connectionist temporal localization (CTL). These losses allow the streaming capability on SLU, enabling the prediction of semantics based on incoming speech. In our first experiments, we showed that our model can achieve accuracy as high as 99.26 \% on non-streaming settings while predicting intent and slot. For streaming scenarios, the proposed model can achieve accuracy of 98.97 \% for CTC and 98.78 \% for CTL on single label prediction. We also showed that our model is able to perform sequential labeling without compromising performance when it is presented with multiple utterances and targets during training and accuracy is as high as 95.69 \% for CTC and 95.28 \% for CTL on two label prediction. As future work, we plan to investigate the capability of our model to precisely locate and decode the semantic span within an utterance.

\newpage

\bibliographystyle{IEEEtran}

\bibliography{mybib}


\end{document}